\documentclass[journal,transmag]{IEEEtran}

\usepackage{cite}

\ifCLASSINFOpdf
   \usepackage[pdftex]{graphicx}

\else

   \usepackage[dvips]{graphicx}

\fi

\usepackage{amsmath}

\usepackage{algorithmic}

\usepackage{array}

\ifCLASSOPTIONcompsoc
  \usepackage[caption=false,font=normalsize,labelfont=sf,textfont=sf]{subfig}
\else
  \usepackage[caption=false,font=footnotesize]{subfig}
\fi

\usepackage{stfloats}

\usepackage{url}

\usepackage{breqn}
\usepackage{amsmath,amssymb}
\usepackage{multirow}
\usepackage{pdflscape}
\usepackage{graphicx,booktabs,array}
\usepackage{multicol}
\usepackage{float}
\usepackage{gensymb}
\usepackage{mhchem}
\usepackage{lipsum}
\usepackage{caption}%% To get \captionof
\usepackage{cite}
\usepackage{array}

\usepackage{breqn}

\usepackage{hyperref}
\hypersetup{
    colorlinks=true,
    linkcolor=blue,
    filecolor=magenta,      
    urlcolor=cyan,
}

\usepackage{xcolor,colortbl}
\usepackage{color,soul}

\begin{document}

\title{Combo Loss: Handling Input and Output Imbalance in Multi-Organ Segmentation}


\author{\IEEEauthorblockN{Saeid Asgari Taghanaki\IEEEauthorrefmark{1,2},
Yefeng Zheng\IEEEauthorrefmark{2},~\IEEEmembership{Senior Member,~IEEE},
S. Kevin Zhou\IEEEauthorrefmark{2},~\IEEEmembership{Senior Member,~IEEE}, \\
Bogdan Georgescu\IEEEauthorrefmark{2},~\IEEEmembership{Senior Member,~IEEE},
Puneet Sharma\IEEEauthorrefmark{2},
Daguang Xu\IEEEauthorrefmark{2}, \\
Dorin Comaniciu\IEEEauthorrefmark{2},~\IEEEmembership{Fellow,~IEEE}, and
Ghassan Hamarneh\IEEEauthorrefmark{1},~\IEEEmembership{Senior Member,~IEEE}}

\IEEEauthorblockA{\IEEEauthorrefmark{1}Medical Image Analysis Lab, School of Computing Science, Simon Fraser University, Canada}
\IEEEauthorblockA{\IEEEauthorrefmark{2}Medical Imaging Technologies, Siemens Healthineers, Princeton, NJ, USA}
 % stops an unwanted space
\thanks{Manuscript received \_ ; revised \_. 
Corresponding author: SA Taghanaki (email:sasgarit@sfu.ca).}}

\markboth{}%
{Shell \MakeLowercase{\textit{et al.}}: Bare Demo of IEEEtran.cls for IEEE Transactions on Magnetics Journals}

\IEEEtitleabstractindextext{%
\begin{abstract}
Simultaneous segmentation of multiple organs from different medical imaging modalities is a crucial task as it can be utilized for computer-aided diagnosis, computer-assisted surgery, and therapy planning. Thanks to the recent advances in deep learning, several deep neural networks for medical image segmentation have been introduced successfully for this purpose. In this paper, we focus on learning a deep multi-organ segmentation network that labels voxels. In particular, we examine the critical choice of a  loss function in order to handle the notorious imbalance problem that plagues both the input and output of a learning model. The input imbalance refers to the class-imbalance in the input training samples (i.e., small foreground objects embedded in an abundance of background voxels, as well as organs of varying sizes). The output imbalance refers to the imbalance between the false positives and false negatives of the inference model. In order to tackle both types of imbalance during training and inference, we introduce a new curriculum learning based loss function. Specifically, we leverage Dice similarity coefficient to deter model parameters from being held at bad local minima and at the same time gradually learn better model parameters by penalizing for false positives/negatives using a cross entropy term. We evaluated the proposed loss function on three datasets: whole body positron emission tomography (PET) scans with 5 target organs, magnetic resonance imaging (MRI) prostate scans, and ultrasound echocardigraphy images with a single target organ i.e., left  ventricular. We show that a simple network architecture with the proposed integrative loss function can outperform state-of-the-art methods and results of the competing methods can be improved when our proposed loss is used.
\end{abstract}

}

\maketitle

\IEEEdisplaynontitleabstractindextext

\IEEEpeerreviewmaketitle

\section{Introduction}

\IEEEPARstart{O}{rgan} segmentation is an important processing step in medical image analysis, e.g., for image guided interventions, radiotherapy, or improved radiological diagnostics. A plethora of single/multi-organ segmentation methods including machine/deep learning approaches has been introduced in the literature for different medical imaging modalities, e.g., magnetic resonance imaging, and positron emission tomography (PET). %We summarize the recent machine/deep learning-based organ segmentation works in Table~\ref{table1}.

More recently, deep learning based medical image segmentation approaches have gained great popularity~\cite{yuan2017automatic, litjens2017survey,baumgartner2017exploration,milletari2017hough,wang2017deepigeos,bentaieb2016topology}. Several deep convolutional segmentation models in the form of encoder-decoder networks have been proposed for both medical and non-medical images to learn features and classify/segment images simultaneously in an end-to-end manner e.g., 2D U-Net~\cite{ronneberger2015u}, 3D U-Net~\cite{cciccek20163d}, 3D V-Net~\cite{milletari2016v}, 2D SegNet~\cite{badrinarayanan2017segnet}. These models with/without modifications have been widely applied to both binary and multi-class medical image segmentation problems. 

%Specifically, for organ segmentation, Cha et al.~\cite{cha2016urinary} applied a patch based CNN method to label each center pixel in a patch to segment bladder from CT scans. However, using only local patches to label center pixels is prone to both false negative (i.e. failure to segment multiple organs with similar intensities/appearances) and false positive errors (small island-like erroneous segmented regions). Also, the densely extracted patches increase the computational cost and running time. 

When performing segmentation especially using deep networks, one has to cope with two types of imbalance issues:

a) Input imbalance or inter-class-imbalance during training, i.e., much fewer foreground pixels/voxels relative to the large number of background voxels in binary segmentation, and smaller objects/classes in a multi-class segmentation relative to other larger objects/classes and the background. Therefore, classes with more observations (i.e., voxels) overshadow the minority classes. 

b) Output imbalance. During inference, it is unavoidable to have false positives and false negatives. False positives are the background voxels (or other objects in the case of multi-class) that are wrongly labeled as the target object. False negatives refer to the voxels of a target object that are erroneously labeled as background or, in the case of multi-organ segmentation, mislabelled as another organ. Clearly, eliminating both false positives and false negatives is the ultimate ideal. However, in practical systems, one increases as the other decreases. For certain applications, reducing the false positive (FP) rate is more important than reducing the false negative (FN) rate or vice versa. The following are example cases where FP should be penalized more: to handle missing organs and to prevent a model from segmenting normal active regions in PET image segmentation (i.e., relatively high intensity regions compared to background which are considered as neither lesion nor organ). PET organ segmentation is useful when a corresponding computed tomography (CT) image is not available to help with detecting organs. Even if a corresponding CT image is available, usually PET and CT need to be registered. In contrast, for some other applications false negatives should be penalized more, e.g., in ultrasound image segmentation where the boundaries of organs are not very clear, target regions might be under-segmented or in magnetic resonance imaging (MRI) segmentation, small spaces of unsegmented regions within a segmented area might be produced. However, conventional loss functions lack a systematic way of controlling the trade-off between false positive and false negative rates. 

A key step when training deep networks on imbalanced data is to properly formulate a loss function. U-Net (both 2D and 3D) and 2D SegNet minimize cross entropy loss to mimic ground truth segmentation masks for an input image while 3D V-Net applies a Dice based loss function.

Cross entropy is commonly used as a loss function in deep learning. Although it can potentially control output imbalance i.e., false positives and false negatives, it has sub-optimal performance when segmenting highly input class-imbalanced images~\cite{badrinarayanan2017segnet}. There are several ways of handling input imbalance in general classification tasks, e.g., random over/under sampling, synthetic minority over-sampling technique (SMOTE)~\cite{chawla2002smote}. Similar to SMOTE, the threshold calibration method, introduced by Pozzolo et al.~\cite{dal2015calibrating}, operates at the data-level, i.e., it requires the data to be undersampled first. However, this cannot be used when the input is an image and we deal with classifying pixels/voxels (i.e., segmentation). To be specific, it would be meaningless to undersample an image by removing only some of its majority class (e.g., background) pixels/voxels in the case of using full-volumes. Although in patch based approaches, patches can be selected in way to handle the imbalance during training, they do not encode full contextual information and the choice of patch size is not straightforward. Therefore, several different techniques such as weighted cross entropy~\cite{sudre2017generalised}, median frequency balancing as used in 2D SegNet~\cite{badrinarayanan2017segnet}, the Dice optimization function as used in the 3D V-Net method~\cite{milletari2016v}, and a focal loss function~\cite{lin2017focal} have been proposed.

Among all methods introduced for tackling the input-imbalance problem, the Dice based loss function has shown better performance for binary-class segmentation problems~\cite{sudre2017generalised}. However, the ability  of the Dice loss function to control the trade-off between false positives and false negatives (i.e., output imbalance) has not been explored in previous works. Controlling the trade-off is not a trivial issue for some types of medical images and it is not easily handled by a classical Dice optimization function. 

Table~\ref{table1_1}, lists previous works that used different loss functions to cope with input/output imbalance. As reported in the table, none of the current loss functions are able to explicitly handle both input and output imbalance. Some other works attempted to enhance output imbalance in the segmented images using post processing techniques, e.g., Hu et al.,~\cite{hu2017automatic} applied an energy based refinement step to improve the CNN segmentation results. Similarly, Gibson et al.,~\cite{gibson2017towards} applied a threshold based refinement step to cope with false positives produced by their convolutional neural network (CNN) based organ segmentation model. Yang et al.~\cite{yang2017automatic} also applied a post processing step to reduce both false positives and false negatives in segmented images. In this paper, we leverage both the cross entropy and the Dice optimization functions to define a new loss function that handles both of the aforementioned input and output imbalance types by using global spatial information driven by Dice term and explicitly and gradually enforcing the trade-off between FNs and FPs by cross entropy term. 

\begin{table}
\caption Applied loss functions in the existing deep models for handling imbalance; In-Imb and Out-Imb refer to input imbalance and output imbalance, respectively
\label{table1_1}
\begin{tabular}{llcc}
\hline
Method               & Loss function                   & In-Imb          & Out-Imb          \\ \hline
2D U-Net~\cite{ronneberger2015u}     & Weighted cross entropy    &                           & \checkmark \\
3D U-Net~\cite{cciccek20163d}       & Weighted cross entropy    &                           & \checkmark \\
3D V-net~\cite{milletari2016v}       & Dice                      & \checkmark &                           \\
Brosch et. al~\cite{brosch2015deep} & Sensitivity + specificity & \checkmark &                           \\
Sudre et. al~\cite{sudre2017generalised}  & Generalized/weighted Dice & \checkmark &                    \\
Berman et. al~\cite{matthew2018lovasz} & Jaccard/IoU               & \checkmark &                           \\
Lin et. al~\cite{lin2017focal} & Focal               &  &              \checkmark              \\
\textbf{Proposed}    & Combo                     & \checkmark & \checkmark \\ \hline
\end{tabular}
\end{table}

In this paper, we make the following contributions: 
a) We introduce a curriculum learning based loss function to handle input and output imbalance (in algorithmic-level) in segmentation problems. 
b) Our proposed loss improves previous deep models namely 3D U-Net, 3D V-Net, and our extended version of 2D SegNet i.e., 3D SegNet in both training and testing accuracy for single and multi-organ segmentation from different modalities.
c) The proposed loss function, by controlling the trade-off between false positives and negatives, is able to handle missing organs i.e., by penalizing the false positives more.
d) We extend 3D U-Net and 3D V-Net from binary to multiclass segmentation models.
e) We introduce the first deep volumetric multi-organ semantic model which simultaneously segments and classifies multiple organs from whole body 3D PET scans. 

\section{Method}

Given a medical image volume, the goal is to predict the class of each voxel by assigning an activation value $p(x)\in [0,1]$ to each voxel x. We adopt a deep learning technique to learn a prediction model $\Phi(x; \theta): x \rightarrow p(x)$, where $\theta$ denotes the model parameters and $p_{i}$ is activation value for organ/class $i$. 

\textbf{Cross Entropy Loss Function.} For multi-class problems, the cross entropy loss can be computed as $C=\sum_{x}\sum_{i} \ t_{i} \ \ln (p_{i(x)})$ where $p$ is the predicted probability mass function (PMF), which assigns a probability/activation value to each class for each voxel, and $t$ is the one-hot encoded target (or ground truth) PMF, where the index $i$ iterates over  the number of organs and $x$ over  the number of the samples (i.e., voxels). $C$ can be computed as a sum of several binary cross entropy terms, which for some multi-class problems, as in this paper, makes it possible to have control over false positives/negatives.  In the case of binary classification, $C$ can be rewritten as $\sum_{x}\ t_{i} \ \ln (p_{i}) + (1-t_{i}) \ \ln (1-p_{i})$. The term $(1-t_{i}) \ \ln (1-p_{i})$ penalizes false positives as it is zero when the prediction is correct. The binary formulation can also be extended and used for multi-class problems as $\frac{1}{N}\sum_{i=1}^{N}t_{i} \ \ln (p_{i}) + (1-t_{i}) \ \ln (1-p_{i})$ where $N=number\ of\ classes \times \ number \ of samples$. Therefore, the output is an average of multiple binary cross entropies.

\textbf{Dice Optimization Function.} The Dice function is a widely used metric for evaluating image segmentation accuracy, which can be written in forms of $Dice=True\ positives\ / \ (number\ of\ positives\ +\ number\ of\ false\ positives)$ or $Dice=2\times TP/(FN+(2\times TP)+FP)$. It can also be rewritten as a weighted function to generalize into multi-class problems~\cite{sudre2017generalised}. However, when it is used as an optimization/loss function, it is not possible to control the penalization of either FPs or FNs separately or their trade-off in the above formulations. In the binary case, the generalized/weighted Dice loss function~\cite{sudre2017generalised} is written as \\ $GDL = 1\left ( 2\left ( \sum_{l=1}^{2} wl\sum_{n}r_{ln}p_{ln} \right ) / \left (\sum_{l=1}^{2}wl\sum_{n}r_{ln}+p_{ln} \right ) \right )$ where $R$ and $P$ are the reference foreground segmentation with voxel values $r_{n}$ and predicted segmentation with voxel values $p_{n}$. However, similar to the original Dice, in this formulation it is not possible to explicitly control the trade-off between FPs and FNs. Moreover, the GDL formulation requires the whole volume to produce meaningful weights (i.e., $wl$), but in most cases, because of limited GPU memory, the segmentation should be performed on sub-volumes. It is also possible to use weighted version of Dice also known as F$_\beta$ score~\cite{hashemi2018asymmetric} to control the trade-off between precision and recall. However, in case of using Dice (F1 or its weighted version F$_\beta$) or GDL with sigmoid activation function in output layer of the network to model probabilities, the derivative of the loss in the back-propagation with respect to a specific weight $w_{jk}$ in layer $L$ looks like:

\begin{equation}
    \frac{\partial Dice}{\partial w_{jk}^{L}} = \frac{1}{n}\sum_x a_{k}^{L-1}\left ( Dice\left ( a_{j}^{L}, y\right ) \right )' \sigma'\left ( z_{j}^{L} \right )
\end{equation}

\noindent where $\sigma'\left ( z_{j}^{L} \right )$ is the derivative of the sigmoid activation function. When a neuron has a value close to 0 or 1, the gradient of the sigmoid is very small. As a result, the gradient of the whole cost function with respect to $w_{jk}$ will become very small. Such a saturated neuron will change its weights very slowly. Note that in equation above Dice (F1) can be replaced by F$_\beta$ or GDL. However, in case of using cross entropy the gradient is computed as

\begin{equation}
    \frac{\partial CE}{\partial w_{jk}^{L}} = \frac{1}{n}\sum_x a_{k}^{L-1}\left ( \sigma\left ( z_{j}^{L} \right )-y \right )
\end{equation}

\noindent Here gradient is not affected by $\sigma'\left ( z_{j}^{L} \right )$ anymore, so the gradient only depends on the neuron’s output, the target $y$ and the neuron’s input $a_{k}^{L-1}$. This avoids learning slow-down and helps with the vanishing gradient problem from which deep neural networks suffer. 

%\begin{multline}
    %GDL = 1-2\left ( \left ( \sum_{l=1}^{2} wl\sum_{n}r_{ln}p_{ln} \right ) / \left (\sum_{l=1}^{2}wl\sum_{n}r_{ln}+p_{ln} \right ) \right )
%\end{multline}

\textbf{Combo Loss.} To leverage the Dice function that handles the input class-imbalance problem, i.e., segmenting a small foreground from a large context/background, while at the same time controlling the trade-off between $FP$ and $FN$ and enforcing a smooth training using cross entropy as discussed above, we introduce our loss $L$ as a weighted sum of two terms: A Dice loss and a modified cross entropy to encode curriculum learning, and is written as:

%\begin{equation}
%\begin{split}
%L= \alpha \sum_{i=1}^{K} \left ( \frac{2\sum_{i=1}^{N} p_i \ %t_i}{\sum_{i=1}^{N} p_i  + \sum_{i=1}^{N} t_i}\right ) + \left ( %1-\alpha \right )  \\ \times \left ( \frac{1}{N} \sum_{i=1}^{N} %\beta \left ( t_i - \ln \ p_i \right  ) + \ \left ( 1-\beta %\right ) \left [ \left ( 1-t_i\right ) \ln \left ( 1-p_i \right %)\right ]\right )
%\end{split}
%\end{equation}

\begin{multline}
    L= \alpha \biggl(-\frac{1}{N} \sum_{i=1}^{N} \beta \left(t_i \ln  p_i \right) \\ + \left(1-\beta \right) \left[\left(1-t_i\right) \ln \left(1-p_i \right)\right]\biggr) \\ - \left ( 1-\alpha \right) \sum_{i=1}^{K} \left ( \frac{2\sum_{i=1}^{N} p_i  t_i + S}{\sum_{i=1}^{N} p_i  + \sum_{i=1}^{N} t_i + S} \right)
\end{multline}

%\begin{equation} 
%\begin{split}
%L & = \alph \left( -\frac{1}{N} \sum_{i=1}^{N} \beta \left(t_i - \ln  p_i \right) \\
% & + \left(1-\beta \right) \left[\left(1-t_i\right) \ln \left(1-p_i \right)\right]\right) \\
% & - \left ( 1-\alpha \right) \sum_{i=1}^{K} \left ( %\frac{2\sum_{i=1}^{N} p_i  t_i}{\sum_{i=1}^{N} p_i  + %\sum_{i=1}^{N} t_i}\right)
%\end{split}
%\end{equation}

\noindent where $\alpha$ controls the amount of Dice term contribution in the loss function $L$, and $\beta \in [0,1]$ controls the level of model penalization for false positives/negatives: when $\beta$ is set to a value smaller than 0.5, $FP$ are penalized more than $FN$ as the term $(1-t_{i}) \ \ln \ (1-p_{i})$ is weighted more heavily, and vice versa. In our implementation, to prevent division by zero, we perform add-one smoothing (a specific instance of the additive/Laplace/Lidstone smoothing)~\cite{russell2016artificial}, i.e. we add unity constant $S$ to both the denominator and numerator of the Dice term. Although the proposed loss seems to be simply combining two different loss functions, we deliberately chose the binary version of the cross entropy to enable us to explicitly enforce a the intended trade-off between false positives and negatives using the parameter $\beta$ (equation 1) and, at the same time, keep the model parameters out of bad local minima via the global spatial information provided by Dice term.

After sigmoid normalization over all the channels (i.e., classes) in the last layer, the Combo loss function is computed using the flattened volumes (one-hot multi-label encoding for both the predicted and ground truth volumes containing several objects) of size $W\times H \times D\times C$ where $W$, $H$, $D$, and $C$ refer to width, height, depth, and number of channels\/classes. This strategy makes it simple to generalize to multi-class segmentation hence directly controlling FPs and FNs over entire volume.

\textbf{Model Parameter Optimization}. To optimize the  model parameters $\theta$ to minimize the loss, we use error back propagation, which relies on the chain rule. We calculate the gradient of $L$ with respect to $p_j$, i.e., $dL/dp_j$, 

%\begin{equation}
%\begin{split}
% \frac{\partial L}{\partial p_j} = 2\alpha \sum_{i=1}^{N}\frac{t_i\left ( \sum_{i=1}^{N} p_i  +\sum_{i=1}^{N} t_i \right )-p_j \left ( \sum_{i=1}^{N} p_i t_i \right )}{\left ( \sum_{i=1}^{N} p_i + \sum_{i=1}^{N} t_i \right )^{2}} + \\ \left ( 1-\alpha  \right ) \times \left ( \frac{1}{N} \sum_{i=1}^{N} \beta \left ( \frac{t_i}{p_i} \right ) +\left ( 1-\beta  \right ) \left ( -\frac{1-t_i}{1-p_i} \right )\right ).
%\end{split}
%\end{equation}

\begin{multline}
   \frac{\partial L}{\partial p_j} =  2\alpha \times \left (-\frac{1}{N} \sum_{i=1}^{N} \beta \left ( \frac{t_i}{p_i} \right ) +\left ( 1-\beta  \right ) \left ( -\frac{1-t_i}{1-p_i} \right )\right ) - \\
    \left ( 1-\alpha  \right ) \sum_{i=1}^{N}\frac{t_i\left ( \sum_{i=1}^{N} p_i  +\sum_{i=1}^{N} t_i \right )-p_j \left ( \sum_{i=1}^{N} p_i t_i \right )}{\left ( \sum_{i=1}^{N} p_i + \sum_{i=1}^{N} t_i \right )^{2}}
\end{multline}

\noindent Then we calculate how the changes in the model parameters in the last layer of the deep architecture affect the predicted $p_j$, and so on.

\begin{table}
\centering
\caption{Comparison between our proposed method and state of the art deep learning segmentation methods. Up\_S refer to up-sampling type.}
\label{table2}
{\renewcommand{\arraystretch}{1.1}%

\begin{tabular}{lllll}
\hline
Network  & SC & Loss          & Up\_S                       & Params \\ \hline
2D SegNet~\cite{badrinarayanan2017segnet}   & No                  & Cross entropy & Specific  & 16,375,169 \\
3D U-Net~\cite{cciccek20163d}    & Yes                 & Cross entropy & Regular                          & 12,226,243 \\
3D V-Net~\cite{milletari2016v}    & Yes                 & Dice          & Regular                          & 84,938,241 \\
Proposed & No                  & Proposed      & Regular                          & 13,997,827 \\ \hline
\end{tabular}}
\end{table}

\begin{figure}
\centering
\includegraphics[scale=.7]{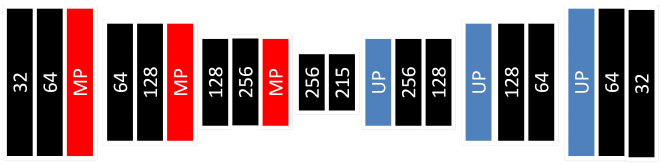}
\caption {The applied network architecture. Black: convolution $3\times3\times3$, MP: max-pooling $2\times2\times2$, UP: up-sampling $2\times2\times2$. The values inside the boxes show number of channels.}
\label{figure2}
\end{figure}

\textbf{Deep Model Architecture.} We use the deep architecture shown in Fig.~\ref{figure2}. This architecture departs from existing architectures like 3D U-Net, 3D U-Net, and 2D SegNet as listed in Table~\ref{table2}. We adopt this simple network to show that the improvement in results is not attributed to some elaborate architecture  and to validate our hypothesis that, even with a simple shallower architecture as long as a proper loss function is used, it is possible outperform more complex architectures e.g., networks with skip connections~\cite{ronneberger2015u,cciccek20163d,milletari2016v} or specific up-sampling~\cite{badrinarayanan2017segnet}.

\textbf{Training.} For multi-organ segmentation from whole body PET images, as the volumes are too large to fit into memory, we extract random sub-volumes from each whole body scan to train a model. Each sub-volume could include voxels belonging to $n \in \{0,1,...,K\}$ organs, with $n=0$ indicating a sub-volume including only background. However, for binary segmentation i.e., 3D ultrasound and MRI datasets, we train using the entire volumes. On test data (only for PET), we apply a volumetric sliding window (with stride), i.e., a volumetric field of view $V$ is partitioned into smaller sub-volumes $\{v_1,v_2,... v_n\}$, where the size of $v_i$ is the same as that of the training sub-volumes. Along any of the dimensions, the stride would be at least 1 voxel and at most the size of the sub-volume in that dimension. Larger strides speed up the computation at the expense of coarser spatial predictions. Let $v_i$ be a subvolume with activation $a_{v_i}$, $V_x$ be the set of subvolumes that include $x$, $A_{V_x}$ be the set of corresponding activation values. $T(A_{V_x})$ is the set of indicator variables whose value is $1$ if the activation is larger than t, and 0 otherwise, where t is a threshold value. Then, the the label assigned to voxel $x$ is given by: $f\left ( x \right ) = \left\{\begin{matrix}
0, & \text{if} \quad \text{max}(T(A_{V_x})) = 0\\ 
1,  & \text{otherwise}
\end{matrix}\right.$. In other words, a single voxel $x$ may reside within multiple overlapping subvolumes; if the activation of any these subvolumes is larger than threshold $T$, then $x$ is assigned $1$, and $0$ otherwise.

\section{Implementation details}
\noindent a) PET multi-organ segmentation: For training the PET multi-organ segmentation network, from each training image, we extract 100 randomly positioned $80\times80\times80$-voxel sub-volumes per organ (5 organs in total: brain, heart, left kidney, right kidney, and bladder) and another 100 for negative background sub-volumes. Therefore, we train all the models with $\sim600\times$ number of training volumes. In test, the striding size was set to $20\times20\times20$. PET volumes size varied from $128\times128\times\sim200$ to $128\times128\times\sim500$. We train and test all the models using two Titan-X GPUs in parallel each with batch-size 1.

\noindent b) Ultrasound echocardiography and prostate MRI segmnetation: We train and test all the models on these datasets with whole-volume images (i.e., not sub-volumes) of size $128\times 128\times128$ and $48\times 256\times 256$ for ultarsound and MRI datasets, respectively using two M5000 GPUs in parallel each with batch-size 2. As explored by Masters et al.,~\cite{masters2018revisiting}, small mini-batch sizes can provide more up-to-date gradient calculations, which results in more stable and reliable training while reducing over-fitting more compared to larger batch sizes. 

As MRI and ultrasound images are taken from a part of the body, the number of slices per volume are relatively less compared to PET whole body volumes. To prevent sliding window for both training and testing and fitting whole MRI and ultrasound volumes into memory, we slightly resampled MRI and Ultrasound images without losing much information causes by resampling. However, PET volumes should be highly resized in order to be fitted into memory which results in considerable accuracy drop. Therefore, we did not resample PET images.

For all datasets, we initialize our models and competing methods using the method introduced by  and Bengio~\cite{glorot2010understanding} and train them with ADADELTA~\cite{zeiler2012adadelta}, with learning rate of 1, $\rho=0.95$, $\epsilon=1e-08$, and $decay=0$. All the models are coded using TensorFlow. To prevent gradient vanishing/exploding we use batch normalization after each convolution layer~\cite{ioffe2015batch}. It also allows us to use higher learning rate. Similar to how hyperparameters values are selected in deep models, e.g., learning rate and pooling window size, the optimal values for $\alpha$ and $\beta$ were also found by grid search to optimize results on the validation set (i.e., one round of cross-validation). We found that the equal contribution (i.e., $\alpha=0.5$) of Dice and cross-entropy terms gives the best results. However, we found that for the PET data, models need to be penalized more for false positives (i.e., $\beta=0.4$) and for MRI and ultrasound data models need to be penalized more for false negatives (i.e., $\beta=0.6$ for MRI and $\beta=0.7$ for ultrasound images). For the last layer of the proposed method, we applied the sigmoid activation function as it allowed us to compute the loss over only foreground objects (i.e., there is no extra channel for the background class, as softmax function requires) and then normalize the output into the range [0-1]. To obtain the segmentation masks we use threshold of 0.5.

All the models have been trained for a fixed number of epochs and we report the results for the best epoch based on the validation set. Note that for the competing methods we set the hyper-parameters as proposed by the authors of these methods. For fairness and to elucidate the direct effect of the proposed Combo loss, when we replace the original loss functions of the competing methods with Combo loss (Tables~\ref{table3} and~\ref{tableV}), we do not change the original network hyper-parameters.

\section{Datasets}
\begin{figure}[H]
\centering
\includegraphics[scale=.72]{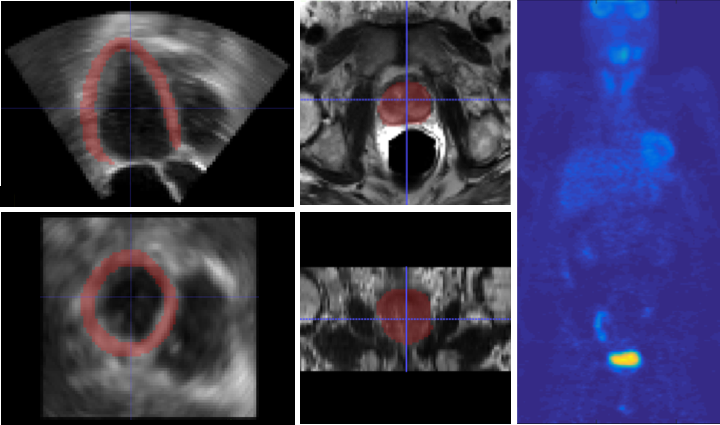}
\caption {Samples of the used datasets. The first column shows the left  ventricular  myocardium (highlighted in red) in ultrasound echocardiography. In the second column, the prostate is highlighted in red in an MRI scan. For MRI and ultrasound samples, the 2 rows show the axial and coronal views. Column 3 shows coronal view of a whole body PET scan.}
\label{figuredata}
\end{figure}

For evaluation, we use three different datasets: a) 58 whole body PET scans of resolution $\sim0.35\times\sim0.35\times\sim3.5-5 \ mm$. We randomly pick 10 whole body volumes for testing and train with the 48 remaining volumes. We normalize the intensity range of our training and testing volumes using the min-max method based on min and max intensity values of the whole training set. Next, in both training and testing, each single sub-volume is also normalized to $[0,1]$ using its min and max before feeding it into network.  b) 958 MRI prostate scans of different resolution which were resampled to voxel size of $1\times1\times3 \ (mm)$. We randomly picked 258 volumes for testing, and train with remaining 700 volumes. c) hl{Ultrasound echocardiography} images of resolution $2\times2\times2 \ (mm)$, used for left ventricular myocardial segmentation, were split into 430 train and 20 test. The datasets were collected internally and from The Cancer Imaging Archive (TCIA) QIN-HEADNECK and ProstateX datasets~\cite{QINheadNeck2015, QINheadNeck2016,ProstateX2017,litjens2014computer,clark2013cancer}. Samples of the three datasets are shown in Fig.~\ref{figuredata}.

\section{Results}
Our evaluation is divided into 2 parts. First, in subsection~\ref{results1}, we compare, both qualitatively and quantitatively, the performance of all the competing methods to the proposed method on the test data, for multi-organ segmentation from PET scans. We test different modification/variants of the proposed loss with the proposed architecture, i.e., cross entropy optimization ($\textrm{P}_{\textrm{CE}}$, weighted cross entropy ($\textrm{P}_{\textrm{WCE}}$), Dice optimization ($\textrm{P}_{\textrm{D}}$), Dice + cross entropy optimization ($\textrm{P}_{\textrm{DCE}}$), and the proposed loss ($\textrm{P}_{\textrm{Combo}}$). DCE refers to simply integrating Dice and traditional cross entropy losses, whereas, Combo refers to combining the weighted version of cross entropy with Dice. Second, in subsection~\ref{results2}, we perform similar experiments to subsection~\ref{results1} for single organ segmentation from two more different modalities, i.e., MRI and ultrasound scans. 

\subsection{Performance of the proposed vs. competing methods on multi-organ PET segmentation}
\label{results1}

\begin{table*}
%\begin{minipage}{\textwidth}
\setlength{\tabcolsep}{6pt}
\centering
\caption{Comparing the performance of competing methods with/without the proposed loss function vs. the proposed method for PET multi-organ segmentation}
\label{table3}
{\renewcommand{\arraystretch}{1.2}%

\begin{tabular}{llcccc}
\hline
&Methods & Jaccard & Dice & FPR & FNR \\ \hline
\multirow{2}{*}{a} &3D U-Net~\cite{cciccek20163d} & $0.55\pm0.16$ & $0.69\pm0.16$ & $0.41\pm0.31$ & $0.30\pm0.15$ \\
&3D U-Net\_Combo & $0.68\pm0.18$ & $0.79\pm0.15$ & $0.18\pm0.09$  & $0.21\pm0.17$ \\ \hline
\multirow{2}{*}{b}&3D V-Net~\cite{milletari2016v} & $0.52\pm0.17$ & $0.67\pm0.16$ & $0.89\pm0.90$   & $0.13\pm0.09$\\
&3D V-Net\_Combo & $0.55\pm0.17$ & $0.70\pm0.15$ & $0.64\pm0.46$  & $0.16\pm0.11$ \\ \hline
\multirow{2}{*}{c}&3D SegNet & $0.41\pm0.20$ & $0.56\pm0.21$ & $0.66\pm0.78$  & $0.32\pm0.28$ \\
&3D SegNet\_Combo & $0.55\pm0.19$ & $0.69\pm0.17$ & $0.37\pm0.38$  & $0.28\pm0.17$\\ \hline
d&Ahmadvand et al.~\cite{ahmadvand2016tumor} & $0.41\pm0.18$ & $0.53\pm0.23$ & $0.35\pm0.77$  & $0.37\pm0.82$\\ \hline
\multirow{2}{*}{e}& $\textrm{P}_{\textrm{CE}}$ & $0.34\pm0.15$ & $0.49\pm0.18$ & $0.67\pm0.42$  & $0.48\pm0.14$\\
& $\textrm{P}_{\textrm{WCE}}$ & $0.45\pm0.20$ & $0.60\pm0.19$ & $0.46\pm0.23$  & $0.37\pm0.22$\\ \hline
\multirow{3}{*}{f}& $\textrm{P}_{\textrm{D}}$ & $0.67\pm0.09$ & $0.80\pm0.07$ & $0.43\pm0.19$  & $0.06\pm0.04$\\
& $\textrm{P}_{\textrm{DCE}}$ & $0.73\pm0.10$ & $0.84\pm0.07$ & $0.09\pm0.06$  & $0.21\pm0.11$ \\
& \textbf{$\textrm{P}_{\textrm{Combo}}$} & $0.73\pm0.13$ & $0.84\pm0.10$ & $0.07\pm0.02$  & $0.22\pm0.14$\\ \hline
\end{tabular}}
%\end{minipage}

\end{table*}

\begin{figure*}
{\renewcommand{\arraystretch}{1.1}%

\resizebox{\textwidth}{!}{\begin{tabular}{*9{c}@{}}
\toprule
 PET (coronal) & GT & Ahmadvand et al.~\cite{ahmadvand2016tumor} & 3D SegNet & 3D U-Net~\cite{cciccek20163d} & 3D V-Net~\cite{milletari2016v} & $\textrm{P}_{\textrm{D}}$  & $\textrm{P}_{\textrm{DCE}}$  & \textbf{$\textrm{P}_{\textrm{Combo}}$}  \\ 
\midrule
 \includegraphics[width=2cm,height=3.5cm]{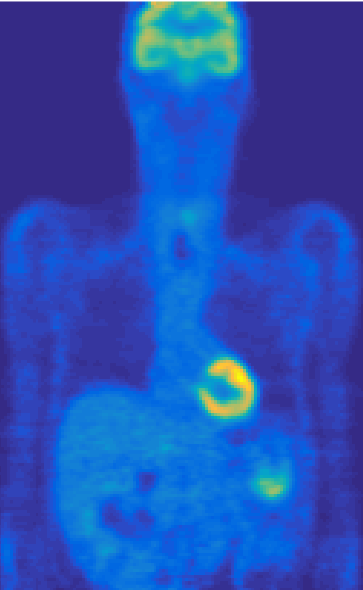}&
\includegraphics[width=2cm,height=4cm]{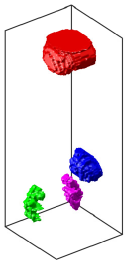}& \includegraphics[width=2cm,height=4cm]{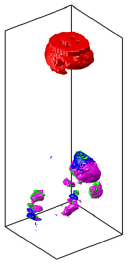}& \includegraphics[width=2cm,height=4cm]{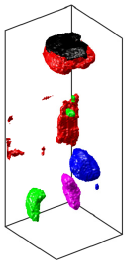}& \includegraphics[width=2cm,height=4cm]{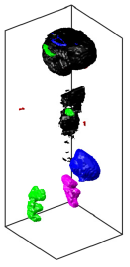}& \includegraphics[width=2cm,height=4cm]{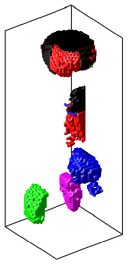}& \includegraphics[width=2cm,height=4cm]{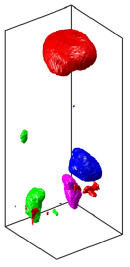}& \includegraphics[width=2cm,height=4cm]{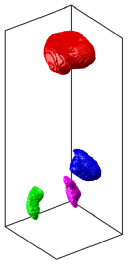}& \includegraphics[width=2cm,height=4cm]{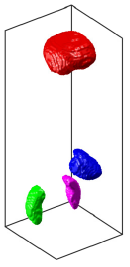} \\ 
\includegraphics[width=2cm,height=3.5cm]{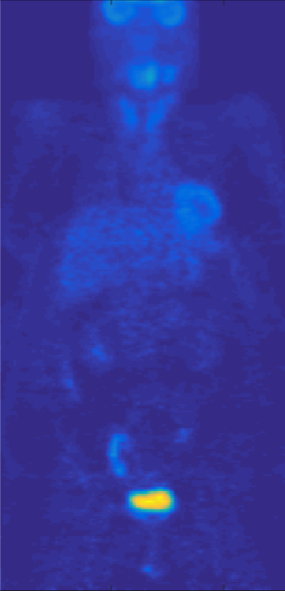}&
 \includegraphics[width=2cm,height=4cm]{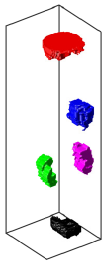}& \includegraphics[width=2cm,height=4cm]{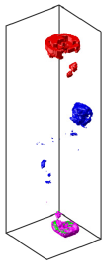}& \includegraphics[width=2cm,height=4cm]{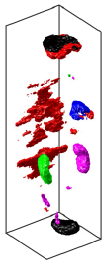}& \includegraphics[width=2cm,height=4cm]{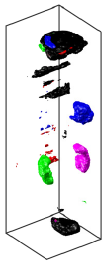}& \includegraphics[width=2cm,height=4cm]{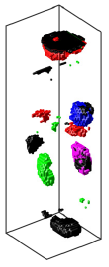}& \includegraphics[width=2cm,height=4cm]{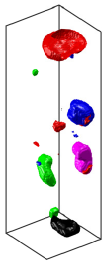}& \includegraphics[width=2cm,height=4cm]{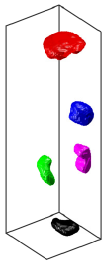}& \includegraphics[width=2cm,height=4cm]{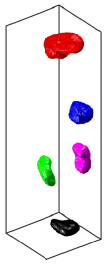} \\ 
 \includegraphics[width=2cm,height=3.5cm]{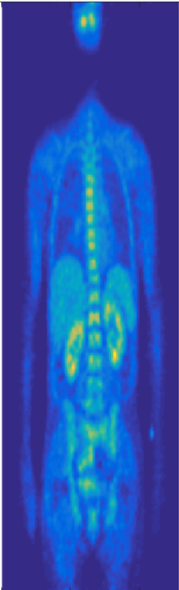}&
 \includegraphics[width=2cm,height=4cm]{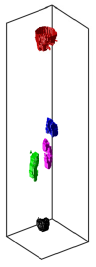}& \includegraphics[width=2cm, height=4cm]{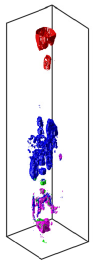}& \includegraphics[width=2cm,height=4cm]{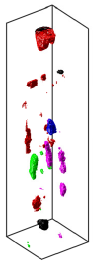}& \includegraphics[width=2cm,height=4cm]{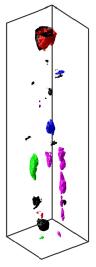}& \includegraphics[width=2cm,height=4cm]{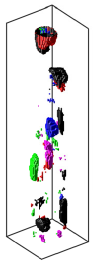}& \includegraphics[width=2cm,height=4cm]{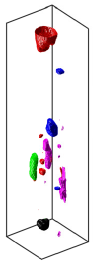}& \includegraphics[width=2cm,height=4cm]{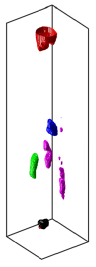}& \includegraphics[width=2cm,height=4cm]{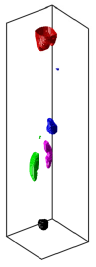} \\
\bottomrule 
\end{tabular}}}

\caption{Comparing multi-organ localization-segmentation-classification results of the proposed vs. competing methods. Each row shows a sample patient data.}
\label{figure3}
\end{figure*} 

We treat the multi-class case as binary, i.e., the one-hot multi-label encoding for the both the predicted and ground truth volumes containing several objects are flattened and the Combo loss is computed. In this case, similar to binary segmentation, balancing the false positives and negatives improves segmentation.
As reported in Table~\ref{table3}, the proposed architecture with proposed loss ($P_{Combo}$) outperforms all competing methods with $57\%\pm24\%$, $38\%\pm18\%$, $86\%\pm5\%$ in Jaccard, Dice and FPR, respectively. 
Comparing rows of section a in Table~\ref{table3}, we note that: Modified 3D U-Net improves with our proposed loss (Combo) relatively by 23.6\%, 14.5\%, 56\%, and 30\% in Jaccard, Dice, FPR, and FNR, respectively. Comparing rows of section  b, we note that: 3D V-Net improves with our proposed loss relatively by 5.8\%, 4.5\%, and 28\%, in Jaccard, Dice, and FPR, respectively. Section c shows that 3D SegNet improves with our proposed loss by relatively 34.1\%, 23.2\%, 44\%, and 12.5\% in Jaccard, Dice, FPR, and FNR, respectively. Comparing $\textrm{P}_{\textrm{CE}}$ vs. $\textrm{P}_{\textrm{WCE}}$ in section e of Table~\ref{table3} shows that WCE helps. Comparing $\textrm{P}_{\textrm{D}}$ vs. $\textrm{P}_{\textrm{Combo}}$ shows that the proposed Combo loss improves the results. 
Although the results and formulation of Dice + original cross entropy (i.e., DCE) and Combo loss are close, it is important to note that, in the Combo loss formulation, we weight the two terms of the original cross entropy so we can enforce the intended trade-off between FP and FN.

As shown in Figure~\ref{table3}, although 3D U-Net, 3D V-Net, and the extended version (3D) of SegNet are able to locate the normal activities (bright areas in the image because of absorbing radio-tracer. The look very similar to abnormalities) and segment them, two issues are visible: a) misclassification of organs: the competing methods were not successful in distinguishing the organs from each other, as sometimes the brain (red) has been labeled as bladder (black); b) the competing methods tend to produce false positives i.e., wrongly labeling some background voxels as an organ (or one organ as another) or missing an organ (false negative).
As shown in the figure, $\textrm{P}_{\textrm{D}}$ still produces false positives, but no misclassification of organs. $\textrm{P}_{\textrm{CE}}$ shows clearer segmentations, however, as we penalize the false positives more with the proposed loss we obtain much clearer outputs (last columns: $\textrm{P}_{\textrm{Combo}}$).   
The performance of the proposed method was evaluated for each specific organ and reported in Table~\ref{table4}.

\begin{figure*}
\centering

{\small
{\renewcommand{\arraystretch}{1.2}%

\begin{tabular}{ccccccc}
\hline
GT & 3D SegNet & 3D SegNet\_Combo & 3D U-Net~\cite{cciccek20163d} & 3D U-Net\_Combo & 3D V-Net~\cite{milletari2016v} & 3D V-Net\_Combo \\ \hline \\

 \includegraphics [height=4cm]{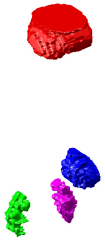} &
 \includegraphics[height=4cm]{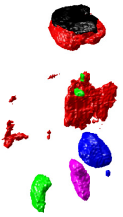} & 
 \includegraphics[height=4cm]{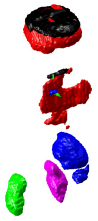} &
 \includegraphics[height=4cm]{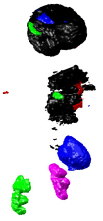} & \includegraphics[height=4cm]{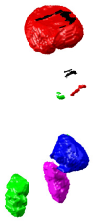} &  \includegraphics[height=4cm]{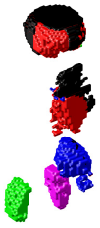} & 
 \includegraphics[height=4cm]{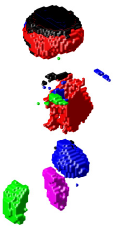} \\ \hline
\end{tabular}}}
\caption{Competing methods' results before and after adding proposed loss function.}
\label{fig4}

\end{figure*}

\begin{figure}
\centering

{\renewcommand{\arraystretch}{1.2}%

\begin{tabular}{lll}
 \includegraphics [scale=.13]{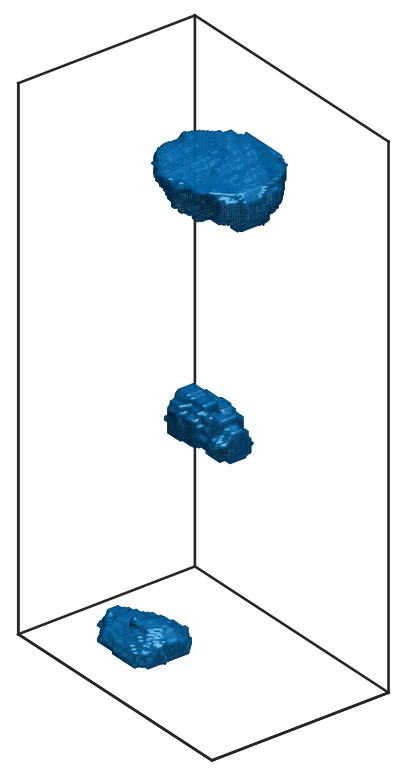}& 
 \includegraphics [scale=.13]{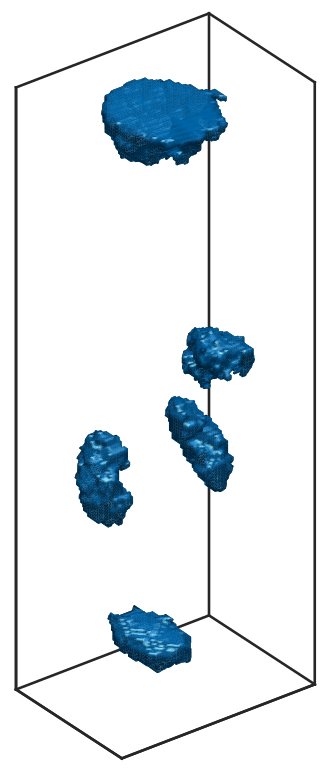} & 
 \includegraphics [scale=.13]{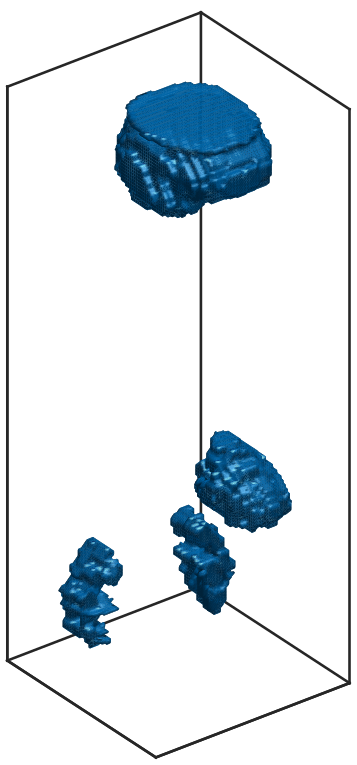} \\
 \includegraphics [scale=.13]{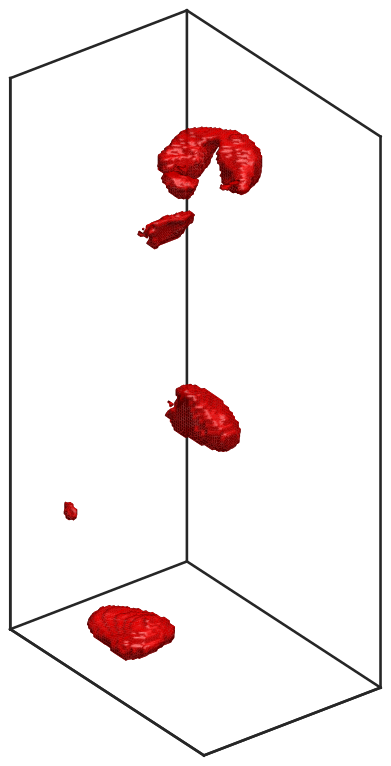}& 
 \includegraphics [scale=.13]{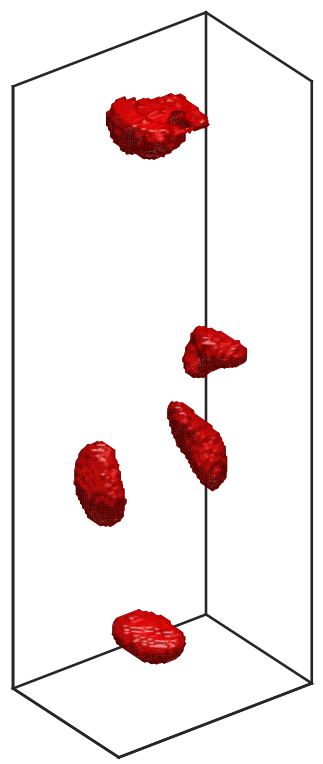} &
 \includegraphics [scale=.13]{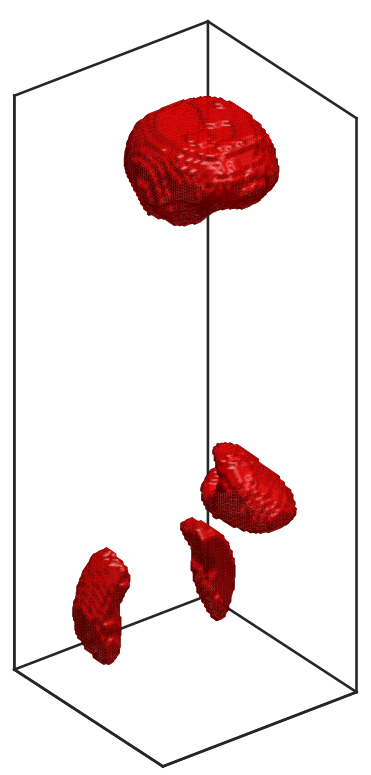}
\end{tabular}}
\caption{Sample segmentations by the proposed method. First row shows ground truth segmentations and second row shows the proposed method results. From left to right: worst (Dice = 0.58), in-between (Dice = 0.76), and best (Dice = 0.91)}
\label{fig5}
\end{figure}

Over all the organs, Dice scores for the proposed method (proposed architecture + Combo loss) ranges from $0.58$ to $0.91$. We show the worst, an in-between and the best results in terms of Dice score in Fig.~\ref{fig5}. Although the left case in the figure seems to be the worst result in terms of Dice score, it is a difficult case with several missing organs. However, the proposed method has been able to handle multiple missing organs to a high extent. Note that some organs c an be physically absent from a patient body, as in renal agenesis or 
radical (complete) nephrectomy, but in PET scans, there might be more "missing" organs (similar to the left case in Fig.~\ref{fig5}) simply because of lack of radiotracer uptake in these organs thus they do not appear in PET. Although, in training, Dice score improvement compared to 3D V-Net is small, as shown in Figure~\ref{fig4}, in test, proposed loss helped 3D V-Net in terms of reducing organ misclassification and false positives. Looking at both Table~\ref{table3} and Fig.~\ref{fig4}, 3D U-Net and 3D SegNet achieved higher performance when incorporating the proposed loss.

\begin{table}
\setlength{\tabcolsep}{4pt}
\centering
\caption{Organ-specific quantitative results of the proposed method for PET dataset. BR, HR, LK, RK, and BL refer to different organs i.e., brain, heart, left kidney, right kidney, and bladder, respectively. The numbers in parentheses in front of each organ show the average percentage number of voxels which belong to that organ in whole volumes.}
\label{table4}
{\renewcommand{\arraystretch}{1.2}%

\begin{tabular}{lcccc}
\hline
             & Jaccard     & Dice        & FPR         & FNR         \\ \hline
BR ($\sim 0.64\%$)        & $0.74\pm0.20$ & $0.83\pm0.16$ & $0.06\pm0.04$ & $0.21\pm0.22$ \\
HR ($\sim 0.14\%$)      & $0.65\pm0.15$ & $0.78\pm0.12$ & $0.11\pm0.07$ & $0.29\pm0.15$ \\
LK ($\sim 0.08\%$) & $0.68\pm0.08$ & $0.81\pm0.06$ & $0.19\pm0.15$ & $0.20\pm0.07$\\
RK ($\sim 0.09\%$) & $0.68\pm0.10$ & $0.81\pm0.07$ & $0.09\pm0.07$ & $0.26\pm0.11$ \\
BL ($\sim 0.09\%$)      & $0.69\pm0.12$ & $0.81\pm0.09$ & $0.04\pm0.05$ & $0.28\pm0.14$ \\ \hline
\end{tabular}}
\end{table}

\subsection{Performance of the proposed vs. competing methods on single organ segmentation from MRI and ultrasound}
\label{results2}

For MRI and ultrasound datasets, we observed that all the methods are more prone to false negatives than false positives, so we weigh more the false negative term of the proposed loss (i.e., increase $\beta$ to 0.9). As reported in Table~\ref{tableV}, similar to results in Section~\ref{results1}, the Combo loss function improved 3D U-Net and 3D V-Net by 4.6\% and 1.13\% in Dice and 43.8\% and 16.7\%in FNR, respectively, for MRI prostate segmentation. Similarly, 3D U-Net and 3D V-Net results were improved by 8.23\% and 3.4\% in Dice and 33.3\% and 16.7\%in FNR, respectively, for ultrasound left ventricular myocardial segmentation.

\begin{table}
\setlength{\tabcolsep}{4pt}
\centering
\caption{Comparing performance of competing methods with/without proposed (\_Combo) loss function vs. proposed method for MRI prostate and ultrasound left ventricular myocardial segmentation. The average percentage of voxels belonging to organ in whole volumes are 1.01\% and 0.6\% for left ventricle and prostate in ultrasound and MRI volumes, respectively.}
\label{tableV}
{\renewcommand{\arraystretch}{1.5}%
\begin{tabular}{llccc}
\hline
                            & Methods                 & Dice                              & FPR                                   & FNR                               \\ \hline
\parbox[t]{1mm}{\multirow{6}{*}{\rotatebox[origin=c]{90}{MRI}}}         & 3D U-Net~\cite{cciccek20163d}                    & $0.87\pm0.07$                     & $0.0004\pm0.0004$           & $0.16\pm0.12$                     \\
& \textbf{3D U-Net\_Combo}  & $0.91\pm0.05$                     & $0.0005\pm0.0005$ & $0.09\pm0.08$  \\
& 3D V-Net~\cite{milletari2016v}               & $0.88\pm0.05$                     & $0.0006\pm0.0004$                     & $0.12\pm0.08$                     \\
& 3D V-Net\_Focal                     & $0.87\pm0.04$                     & $0.0002\pm0.0002$                     & $0.19\pm0.07$                     \\
& 3D V-Net\_Combo              & $0.89\pm0.05$                     & $0.0006\pm0.0005$                     & $0.10\pm0.08$                     \\
 & ProposedArc\_Combo          & $0.90\pm0.04$                     & $0.0007\pm0.0006$                     & $0.08\pm0.07$                     \\ \hline
\parbox[t]{1mm}{\multirow{6}{*}{\rotatebox[origin=c]{90}{Ultrasound}}} & 3D U-Net~\cite{cciccek20163d}  & \multicolumn{1}{l}{$0.85\pm0.05$} & \multicolumn{1}{l}{$0.0020\pm0.0006$}  & \multicolumn{1}{l}{$0.12\pm0.12$} \\
                            & \textbf{3D U-Net\_Combo}   & \multicolumn{1}{l}{$0.92\pm0.05$} & \multicolumn{1}{l}{$0.0007\pm0.0003$} & \multicolumn{1}{l}{$0.08\pm0.09$} \\
                            & 3D U-Net\_Focal   & \multicolumn{1}{l}{$0.88\pm0.11$} & \multicolumn{1}{l}{$0.0004\pm0.0005$} & \multicolumn{1}{l}{$0.17\pm0.15$} \\
                            & 3D V-Net~\cite{milletari2016v} & \multicolumn{1}{l}{$0.84\pm0.04$} & \multicolumn{1}{l}{$0.0020\pm0.0008$}  & \multicolumn{1}{l}{$0.12\pm0.08$} \\
                            & 3D V-Net\_Combo & \multicolumn{1}{l}{$0.87\pm0.03$} & \multicolumn{1}{l}{$0.0020\pm0.0005$}   & \multicolumn{1}{l}{$0.10\pm0.04$} \\
                            & \textbf{ProposedArc\_Combo}  & \multicolumn{1}{l}{$0.92\pm0.05$} & \multicolumn{1}{l}{$0.0006\pm0.0004$} & \multicolumn{1}{l}{$0.09\pm0.10$} \\ \hline
\end{tabular}}
\end{table}

As can also be seen in Table~\ref{tableV}, the proposed loss also helps reduce the variance of the segmentation results. 

\begin{figure}[H]
\centering
\includegraphics[scale=.55]{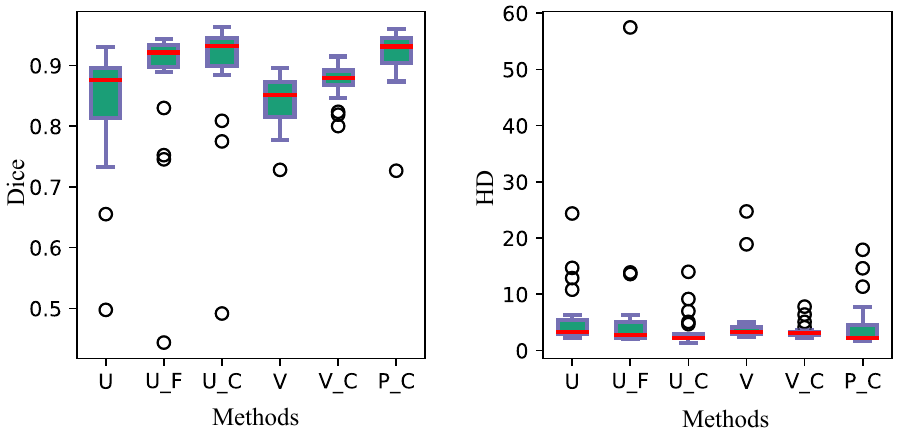}
\caption {Dice and Hausdorff distance (HD) of the competing methods vs. the proposed Combo loss for the ultrasound dataset. In the figure, U, U\_F, U\_C, V, V\_C, and P\_C refer to 3D U-Net, 3D U-Net\_Focal, 3D U-Net\_Cmbo, 3D V-Net, 3D V-Net\_Combo, and ProposedArc\_Combo, respectively. }
\label{hddiceecho}
\end{figure}

We also compared the proposed loss function with the recently introduced Focal loss function~\cite{lin2017focal}. Our integrative loss function outperformed Focal loss after both were applied to different networks (Table~\ref{tableV}). We applied Focal loss to the best performing competing method for each dataset i.e., 3D V-Net for MRI and 3D U-Net for ultrasound dataset. For Focal loss, we tested several different values for $\alpha$ and $\gamma$, but as suggested by the authors we obtained better results with $\alpha=0.25$ and $\gamma=2.0$. Note that there is no correspondence between the alpha used in the Focal loss paper (the weight assigned to the rare class)  and the one we use in Combo loss equation (the weight that controls the contribution of Dice and cross entropy terms). For the MRI dataset, the proposed Combo loss outperformed Focal loss by $2.3\%$ and $47.4\%$ in Dice and FNR, respectively, when both were used in 3D V-Net. For the ultrasound dataset, Combo loss outperformed Focal loss by $10.8\%$ in Dice. In Figure~\ref{hddiceecho}, we plot both Dice and Hausdorff distance (HD) of the Combo loss vs. competing methods. As shown in the figure, the proposed method outperforms the competing methods in terms of Dice score. Comparing both Dice and Hausdorff distance values of the competing methods, after applying Combo loss (i.e., U\_C, and V\_C) in Figure~\ref{hddiceecho}, the range of the values are smaller, i.e., less outliers compared to when they use original loss (i.e., U and V).

Among the competing methods, U-Net applies cross entropy loss while V-Net leverages Dice loss. To show the direct contribution of the Combo loss, we replace the original loss functions in U-Net and V-Net with Combo (Table~\ref{tableV}). As reported in Table~\ref{tableV}, after replacing cross entropy loss of U-Net with Combo loss, the Dice scores improve from 0.87 to 0.91 and 0.85 to 0.92 for MRI and ultrasound datasets, respectively. Similarly, when replacing the Dice loss function of V-Net with the proposed Combo loss, the segmentation results improve from 0.88 to 0.89 and 0.84 to 0.87 for MRI and ultrasound datasets, respectively.

%\hl{In Figure}~\ref{betach}, \hl{we show the different Dice and HD results obtained from different $\beta$ values which controls false positives and negatives. As shown, the best results obtained by $\beta=0.6$ and $\beta=0.7$ for MRI and ultrasound images, respectively. As HD is sensitive to outliers, there are sometimes relatively large values in the HD results (i.e. second column in the figure).}

\begin{figure}[H]
\centering
\includegraphics[scale=.55]{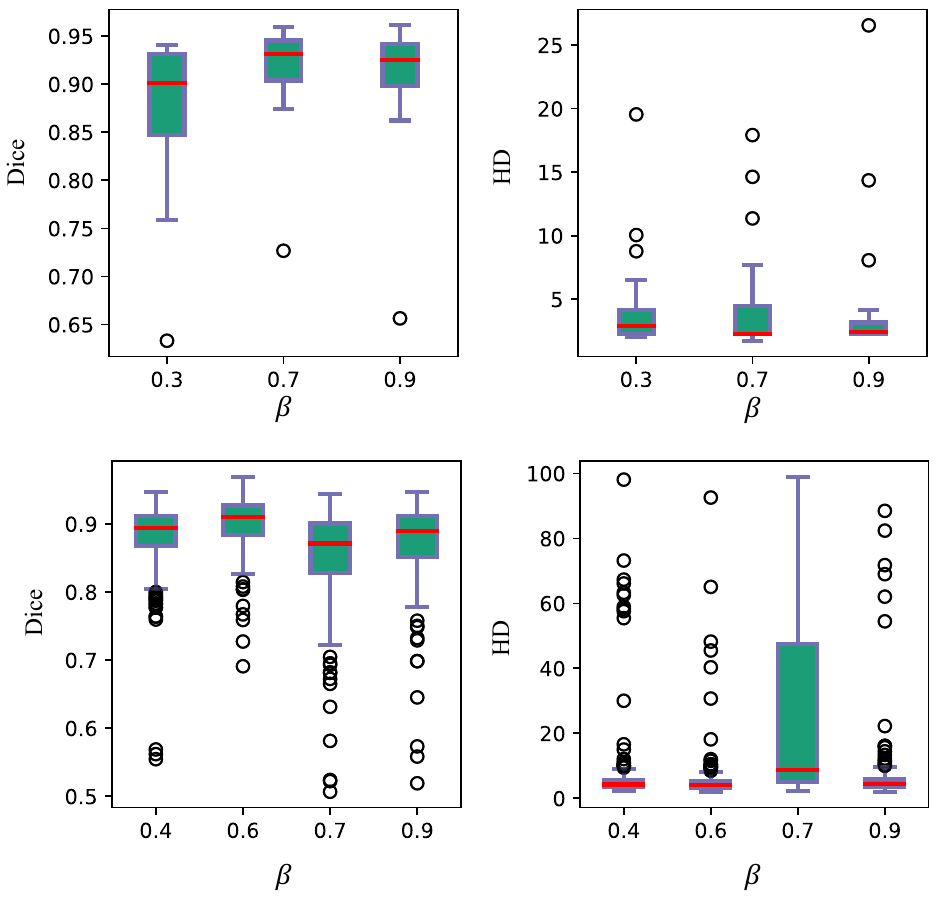}
\caption {Dice and Hausdorff distance (HD) with various parameter $\beta$ values. First and second rows show ultrasound and MRI results, respectively.}
\label{betach}
\end{figure}

Parameter $\alpha$ controls the contribution of Dice and cross entropy terms while parameter $\beta$ in the second term, i.e., cross entropy, controls the trade-off between false positives and negatives. As a key contribution of the paper is providing the means to explicitly control output balance, i.e., false positives and negatives, we tested several different values for parameter beta to see how the final results are affected by $\beta$ and we fix parameter $\alpha$ that controls the trade-off between Dice and cross entropy to 0.5. In Figure~\ref{betach}, we show the different Dice and HD results obtained from different $\beta$ values, which control false positives and false negatives. As expected, we note that the final segmentations are affected by the choice of parameter beta and the best results in terms of higher Dice and lower Hausdorff distance were obtained for $\beta = 0.7$ and $\beta = 0.6$ for ultrasound and MRI datasets, respectively. As HD is sensitive to outliers, there are sometimes relatively large values in the HD results (i.e., second column in the figure)

\section{Conclusion}
In this paper, we proposed a curriculum learning based loss function to handle input/class-imbalance and output imbalance (i.e., enforcing the trade-off between false positives and false negatives). Note that enforcing a desired  trade-off between false positives and false negatives can be seen in Tables~\ref{table3} and~\ref{tableV}). Noting the change in  FPR and FNR values of 3D U-Net, 3D V-Net, and 3D Seg-Net when they apply Combo loss, we see that FPR or FNR is severely decreased when the models are penalized for FP or FN, respectively (for PET data i.e., Table III, the Combo loss penalizes FP and for MRI and ultrasound data i.e., Table V, it penalizes FN). The proposed loss function resulted in improved performance in both multi- and single-organ segmentation from different modalities. The proposed loss function also improved the existing methods in terms of achieving higher Dice and lower false positive and false negative rates. In this work, we applied the proposed loss function to a multi-organ segmentation problem, but it can simply be leveraged for other segmentation tasks as well. The key advantage of the proposed Combo loss is that it enforces a \emph{desired} trade-off between the false positives and negatives (which results in cutting out post-processing) and avoids getting stuck in bad local minima as it leverages Dice term. The Combo loss converges considerably faster than cross entropy loss during training. Similar to Focal loss, our Combo loss also has two parameters that need to be set. In this work, we used cross-validation to set the hyperparameters (including $\alpha$ and $\beta$ of our proposed loss). Future work can explore using Bayesian approaches~\cite{snoek2012practical, murugan2017hyperparameters}.

%\noindent \textbf{Disclaimer:} This feature is based on research and is not commercially available. Due to regulatory reasons its future availability cannot be guaranteed.
\section*{Acknowledgment}
We thank NVIDIA for GPU donation.

\bibliographystyle{IEEEtran}
\bibliography{IEEEexample.bib}
\end{document}